\documentclass[twocolumn]{article}
\usepackage{graphicx} 
\usepackage{amsmath}
\usepackage{amssymb}
\usepackage{hyperref}
\usepackage{geometry}
\usepackage{booktabs} %
\usepackage{amsfonts}
\usepackage{multirow}
\usepackage{amsthm}
\usepackage{enumitem}
\usepackage{dcolumn}
\usepackage{bm}
\usepackage{longtable}
\usepackage{float}
\usepackage{tabularx}   %
\usepackage{epsfig}
\input{epsf}
\usepackage{psfrag}
\usepackage{xcolor}
\usepackage{pdfpages}
\usepackage[normalem]{ulem}
\usepackage{subcaption} %
\usepackage[numbers,sort&compress]{natbib}
\usepackage{authblk}

\newtheorem{definition}{Definition}

\newtheorem{lemma}{Lemma}
\newtheorem{proposition}{Proposition}

\newtheorem{remark}{Remark}
\newtheorem{notation}{Notation}
\newtheorem{assumption}{Assumption}

\usepackage[utf8]{inputenc}
\usepackage{listings}
\usepackage{xcolor}
\usepackage{inconsolata}

\lstdefinestyle{mypython}{
  language=Python,
  basicstyle=\ttfamily\small,
  keywordstyle=\color{blue!70!black},
  stringstyle=\color{orange!70!black},
  commentstyle=\color{green!60!black},
  showstringspaces=false,
  breaklines=true,
  frame=single,
  numbers=left,
  numberstyle=\tiny\color{gray},
  xleftmargin=2em,
  xrightmargin=2em,
  aboveskip=1em,
  belowskip=1em
}
\usepackage{fancyvrb}   %

\author[1]{Zhuo-Yang Song}
\affil[1]{School of Physics, Peking University, Beijing 100871, China}

\begin{document}

\twocolumn[
  \begin{@twocolumnfalse}
    \title{Where to Search: \\Measure the Prior-Structured Search Space of LLM Agents}
    \date{}
    \maketitle
    \begin{abstract}
    The generate-filter-refine (iterative paradigm) based on large language models (LLMs) has achieved progress in reasoning, programming, and program discovery in AI+Science. However, the effectiveness of search depends on where to search, namely, how to encode the domain prior into an operationally structured hypothesis space. To this end, this paper proposes a compact formal theory that describes and measures LLM-assisted iterative search guided by domain priors. We represent an agent as a fuzzy relation operator on inputs and outputs to capture feasible transitions; the agent is thereby constrained by a fixed safety envelope. To describe multi-step reasoning/search, we weight all reachable paths by a single continuation parameter and sum them to obtain a coverage generating function; this induces a measure of reachability difficulty; and it provides a geometric interpretation of search on the graph induced by the safety envelope. We further provide the simplest testable inferences and validate them via two instantiation. This theory offers a workable language and operational tools to measure agents and their search spaces, proposing a systematic formal description of iterative search constructed by LLMs.
    \end{abstract}
    \vspace*{1cm} %
    
  \end{@twocolumnfalse}
]

\section{Introduction}
The generate-filter-refine iterative paradigm centered on large language models (LLMs) is rapidly expanding its application boundary—from reasoning and programming \cite{yao2023reactsynergizingreasoningacting,yao2023treethoughtsdeliberateproblem,NEURIPS2022_9d560961,wang2023selfconsistencyimproveschainthought,Besta_Blach_Kubicek_Gerstenberger_Podstawski_Gianinazzi_Gajda_Lehmann_Niewiadomski_Nyczyk_Hoefler_2024}, to planning and tool use \cite{wang2023planandsolvepromptingimprovingzeroshot,schick2023toolformerlanguagemodelsteach,wang2023voyageropenendedembodiedagent,chen2023programthoughtspromptingdisentangling,pmlr-v202-gao23f,liang2023codepolicieslanguagemodel}, and further to program/function search in AI+Science \cite{mankowitz2023faster,funsearch,alphaevolve,Song:2025pwy,Cao:2025shc,song2025iteratedagentsymbolicregression}. The common structure of such paradigms embeds tasks or hypotheses into an operational space and performs multi-round generation, evaluation, and update on that space. Although this approach has performed well in many cases, its effectiveness is fundamentally constrained by where to search \cite{HOGG19961, 6472238, idofmodel,song2025bridgingdimensionalchasmuncover,song2025iteratedagentsymbolicregression}: that is, how the prior is encoded into the agent-operable space. In practice, agents based on LLMs often do not wander blindly in the original space, but iterate within a smaller semantic space defined by priors and constraints; the geometry and boundary of this space determine efficiency and stability \cite{wolpert2002no}.

Long-horizon tasks raise higher demands for understanding such search. First, safety is the primary constraint: in real systems or sensitive scenarios, LLMs must operate within verifiable and controllable boundaries \cite{altman2021constrained,NEURIPS2022_b1efde53,bai2022constitutionalaiharmlessnessai,christiano2017deep,irving2018aisafetydebate,10.1145/3591300,rebedea2023nemoguardrailstoolkitcontrollable}. Intuitively, this requires formally confining the model within a safety envelope, allowing only constraint-satisfying transitions. Second, complexity requires a systematic characterization of the search process: long-horizon problems often involve combinatorial explosion and sparse rewards; purely heuristic or 0/1 scoring is insufficient to quantify reachability difficulty, compare the coverage capability of different agents, or guide sampling budgets and staged training \cite{HOGG19961,sutton1998reinforcement,10.1145/1553374.1553380}. Therefore, a concise, computable, model-agnostic formal theory is needed: one that unifies safety and reachability under the same set of measures, and provides testable predictions and engineering-usable design principles.

Current practice mostly relies on engineering heuristics (prompt design, filters, scoring functions, temperatures, and sampling budgets), lacking a unified language and quantitative tools for agent-space-search. Concretely, it is difficult to measure, in a comparable way, the trade-offs between reachability and safety across agents, and there is a lack of clear characterization and explanation of long-horizon behavioral features of agents. This theoretical gap may be the key deficiency in moving LLM-driven complex tasks from usable to controllable and measurable.

To address this, the paper proposes a compact formal theory to characterize and measure LLM-assisted iterative search. Specifically, we formalize agents as fuzzy relation operators; in iterative applications we feed outputs back into inputs to form an iterated agent, and introduce a critical parameter as a unified quantification of reachability difficulty. On the directed graph induced by the safety envelope, we discuss geometric features of the search space. To validate the abstract concepts, we provide two minimal instantiation. The instantiation yields evidence consistent with the hypotheses, providing an initial external validation of the formalization.

The significance of this theory is that it establishes a measurement system in which safety and reachability are measured by the same symbols and geometric quantities; this enables operational metrics for several questions—for example, whether an intermediate waypoint can significantly reduce overall difficulty can be localized by the compositional lower bound for coverage (transitivity) of the coverage index. This formalization offers a consistent baseline for comparing agents, designing search strategies, and setting training signals.

The paper is organized as follows. Section 2 presents the formal theory, including the fuzzy relation operator representation of agents, the coverage generating function and critical parameters, geometric quantities and inequalities on the safety envelope-induced graph, and several hypotheses for iterated agents constructed by LLMs. Section 3 provides the majority-vote instantiation and tests the inequalities involving shortest distance and the number of shortest paths, as well as the approximately unidirectional search hypothesis. Section 4 concludes and discusses prospective directions for connecting the proposed measures to evaluation, search policy, and reinforcement learning rewards.

\section{Formal Theory}
\label{sec:formal}

\subsection{Conventions and Objects of Study}

This section introduces the minimal mathematical objects for characterizing the LLM-driven generate-filter-refine process and defines reachability and search geometry using a unified generating-function language.

\begin{notation}[Search space and empty-product convention]
Let $C_1, C_2$ be nonempty sets representing the input space and output space of an agent, respectively. In iterative scenarios, we assume $C_2 \subseteq C_1$ so that outputs can be fed back as inputs for the next step. For any finite product, the empty product is defined to be $1$.
\end{notation}

\begin{definition}[Ideal agent and fuzzy relation operator]
An ideal agent $\mathcal{T}$ is a mapping $f\mapsto \mu_f(\cdot)$, where each $f\in C_1$ is associated with a membership function
\begin{align}
\mu_f : C_2\to [0,1],\qquad g\mapsto \mu_f(g).
\end{align}
This can be equivalently viewed as a fuzzy relation operator $\mathcal{T}(f,g):=\mu_f(g)$ \cite{belohlavek2012fuzzy}.
\end{definition}

\begin{definition}[Crisp idealized agent and safety envelope]
If for all $f\in C_1,\,g\in C_2$, we have $\mu_f(g)\in\{0,1\}$, then $\mathcal{T}$ is called a crisp idealized agent. Fix a crisp idealized agent and denote it by the safety envelope $\mathcal{T}_0$. An ideal agent $\mathcal{T}$ is said to be constrained by $\mathcal{T}_0$ in safety if
\begin{align}
0\le \mathcal{T}(f,g)\le \mathcal{T}_0(f,g),\qquad \forall\, f,g.
\end{align}
In this case, each feasible transition of $\mathcal{T}$ is limited to the reachable edges allowed by $\mathcal{T}_0$, so execution proceeds only within the safety envelope.
\end{definition}

\begin{definition}[Iterated agent and search trajectory]
When $C_2\subseteq C_1$, $\mathcal{T}$ is called an iterated agent. A finite sequence $ST=(f^{(0)},f^{(1)},\dots,f^{(n)})$ is called a search trajectory from $f^{(0)}$ to $f^{(n)}$ (of length $n$) if for all $i=0,\dots,n-1$,
\begin{align}
\mu_{f^{(i)}}\!\big(f^{(i+1)}\big)>0.
\end{align}
\end{definition}

\subsection{Coverage generating function}

To uniformly measure the reachability of iterated agents across problem difficulties, we introduce a coverage generating function based on a continuation parameter without aftereffects.

\begin{definition}[Coverage generating function and continuation parameter]
Let a single parameter $p\in[0,1]$ denote the weight for continuing iteration (the continuation parameter), understood as a scalar weight assigned to trajectory length; it is not a probability but a bookkeeping factor. Thus a trajectory of length $n$ is assigned weight $p^n$. Define the coverage generating function from $f$ to $g$ as
\begin{align}
P_{f,g}(p)
:= \sum_{n=0}^{\infty}\;\sum_{\substack{ST: f^{(0)}=f,\\ f^{(n)}=g}}
\; p^n \prod_{i=0}^{n-1} \mu_{f^{(i)}}\!\big(f^{(i+1)}\big),
\label{eq:coverage_gf}
\end{align}
where for $n=0$ the inner product is the empty product, and this term exists and contributes $1$ if and only if $f=g$.
\end{definition}

\begin{remark}[Operator viewpoint and spectral radius]
If $C_1, C_2$ are countable, let the matrix (kernel) $M$ satisfy $M_{f,g}=\mathcal{T}(f,g)$. Then
\begin{align}
P(p) \;=\; \sum_{n\ge 0} p^n M^n,
\end{align}
whose $(f,g)$ entry is exactly \eqref{eq:coverage_gf}. When $p\,\rho(M)<1$ (with $\rho(M)$ the spectral radius), the series converges in the operator sense and
\begin{align}
P(p) \;=\; (I-pM)^{-1}.
\end{align}
In general, $P_{f,g}(p)$ is a power series (generating function) with nonnegative coefficients, monotone nondecreasing in $p$. The boundary value satisfies $P_{f,g}(0)=\mathbf{1}\{f=g\}$.
\end{remark}

\begin{notation}[Continuation-induced search]
Given an iterated agent $\mathcal{T}$ and parameter $p\in[0,1]$, define the continuation-induced ideal agent
\begin{align}
\mathcal{T}^{(p)}(f,g):=\min\!\big(1,\,P_{f,g}(p)\big),
\end{align}
which can be viewed as an agent formed by multi-round iterative feedback through $\mathcal{T}$. We will refer to $\mathcal{T}^{(p)}$ as the search or the continuation-induced (search) agent. This clipping induces a [0,1]-valued membership, not a probability measure; alternative normalizations are possible, but we adopt unit clipping for threshold analysis.
\end{notation}

\subsection{Geometry under the crisp safety envelope}

On the directed graph induced by the crisp idealized agent (safety envelope), natural geometric quantities can be defined.

\begin{definition}[Generating function under the crisp idealized agent and path counting]
If $\mathcal{T}$ is a crisp idealized agent, then
\begin{align}
P^{\mathrm{ideal}}_{f,g}(p) \;=\; \sum_{n=0}^{\infty} N_n(f,g)\,p^n,
\end{align}
where $N_n(f,g)$ is the number of reachable paths of length $n$ from $f$ to $g$.
\end{definition}

\begin{definition}[Shortest distance]
On the directed graph induced by the crisp idealized agent, define the shortest distance
\begin{align}
d_0(f,g) := \inf\big\{n\in\mathbb{N}:\; N_n(f,g)\ge 1\big\},
\end{align}
and set $d_0(f,g)=+\infty$ if $g$ is unreachable from $f$.
\end{definition}

\begin{lemma}[Shortest distance and low-order terms of the generating function]
If $d_0(f,g)<\infty$, then
\begin{align}
\lim_{p\to 0^+} \frac{P^{\mathrm{ideal}}_{f,g}(p)}{p^{d_0(f,g)}} \;=\; N_{d_0(f,g)}(f,g)\;\in\;\mathbb{N}\setminus\{0\}.
\end{align}
\label{le:low}
\end{lemma}

\begin{remark}[Insufficient search under small continuation parameter]
When the continuation parameter $p$ is small, search is in an insufficient expansion regime (particularly when many edges have low membership or the graph is approximately unidirectional). By Lemma~\ref{le:low}, the shortest-path term dominates the behavior of $P_{f,g}(p)$, so the shortest distance $d_0$ and the corresponding number of shortest paths $N_{d_0}$ control the early reachable set.
\end{remark}

\subsection{Critical parameter and coverage index}

We characterize the reachability difficulty from $f$ to $g$ by the critical value at which the generating function reaches the unit threshold.

\begin{definition}[Coverage index and critical parameter]
Define
\begin{align}
p_c(f,g):=\inf\big\{p\in[0,1]:\; P^{\mathrm{ideal}}_{f,g}(p)\ge 1\big\},
\end{align}
and set $p_c(f,g)=1$ if the set is empty (unreachable). Define the coverage index
\begin{align}
R_c(f,g):=1-p_c(f,g)\in[0,1].
\end{align}
A larger $R_c(f,g)$ indicates reaching unit coverage at smaller weights, i.e., easier to reach. Increasing the number of reachable paths or shortening path length both increase $R_c$.
\end{definition}

\begin{definition}[Intermediate node]
A node $h$ is called an intermediate node for $(f,g)$ if at least one reachable path from $f$ to $g$ passes through $h$.
\end{definition}

\begin{proposition}[Transitivity of the coverage index]
If $h$ is an intermediate node for $(f,g)$, then for all $p\in[0,1]$,
\begin{align}
P^{\mathrm{ideal}}_{f,g}(p) \;\ge\; P^{\mathrm{ideal}}_{f,h}(p)\cdot P^{\mathrm{ideal}}_{h,g}(p).
\end{align}
Therefore,
\begin{align}
p_c(f,g)\;&\le\; \max\!\big(p_c(f,h),\,p_c(h,g)\big),\nonumber\\
R_c(f,g)\;&\ge\; \min\!\big(R_c(f,h),\,R_c(h,g)\big).
\label{eq:chuandi}
\end{align}

If $h$ is not an intermediate node for $(f,g)$, then at least one of $f\to h$ or $h \to g$ is unreachable; in this case Eq.~\ref{eq:chuandi} still holds.
\end{proposition}

\begin{definition}[Epoch and the lower-bound meaning of shortest distance]
An epoch refers to one expansion step that applies the crisp safety envelope $\mathcal{T}_0$ to all outputs from the previous round and performs set-wise deduplication. Clearly, starting at $f$, reaching $g$ requires at least $d_0(f,g)$ epochs.
\end{definition}

\subsection{Threshold hypotheses and testable inequalities}

We provide an empirically common and testable hypothese for LLM-induced approximately unidirectional search.

\begin{assumption}[Approximate strcture of LLM induced agent]
Empirically, for iterated agents constructed by LLMs:
\begin{enumerate}
    \item Closed walks (nonzero-length paths whose start and end coincide) are rare; the crisp envelope is approximately unidirectional, so $P^{\mathrm{ideal}}_{f,g}(p)$ has a much simplier denominator compared to its complete numerator.
    \item Overly long trajectories are relatively rare; equivalently, the generating function is essentially dominated by its low-order terms.
\end{enumerate}
\label{ass:threshold}
\end{assumption}

\section{Instantiation and Experiments}

This section provides two minimal, reproducible instantiation that aligns one-to-one with the formal objects above.

\subsection{Majority-vote walking agent}
On a two-dimensional grid, we construct an ideal agent and its corresponding crisp agent induced by LLM majority vote, and test the observable hypotheses and inferences in Assumption~\ref{ass:threshold}.

To make abstract concepts concrete, we give a minimal construction following objects-mappings-geometry.

\paragraph{Task space and transition syntax:} Consider a two-dimensional grid $G_N := \{0,\dots,N-1\}^2$, hence $C_1=C_2=G_N$. Given a start-target pair $(f,t)\in G_N\times G_N$, we allow unit-step transitions up, down, left, and right, staying within the board as the syntax of feasible local transitions.

\paragraph{Ideal agent induced by LLMs:} For any $f\in G_N$ and fixed target $t$, we query a given LLM ($\mathcal{L}$) under prompts that combine constraints and goals to output the next position $g$; the complete prompt is in Appendix App.~\ref{app:prompt}. This yields an empirical decision distribution $\widehat{P}^{(\mathcal{L},t)}_f(g)$ (approximated via multiple samples) for from $f$ to $g$.

For each $f$, independently sample $m$ times and take the mode $g^\star$ of $\widehat{P}^{(\mathcal{L},t)}_f(g)$. If its frequency exceeds $m/2$, define the agent
    \[
    \mu^{(\mathcal{L},t)}_f(g):=\mathbf{1}\{g=g^\star\};
    \]
    otherwise regard it as no strict majority, i.e., $\mu^{(\mathcal{L,t})}_f(g)=0$.

Aggregate majority-vote results from $n$ different models uniformly to obtain the ideal agent
\[
\mu^{(t)}_f(g)\;:=\;\frac{1}{n}\sum_{\mathcal{L}}\mu^{(\mathcal{L},t)}_f(g)\;\in[0,1].
\]

\paragraph{Crisp agent (safety envelope) and induced graph:} Binarize the support of the ideal agent to obtain a crisp idealized agent
\[
\mu^{0,(t)}_f(g)\;:=\;\mathbf{1}\big\{\mu^{(t)}_f(g)>0\big\}\in\{0,1\},
\]
and define the directed graph $\mathcal{G}_t$: nodes are $G_N$, and if $\mu^{0,(t)}_f(g)=1$ draw an edge $f\to g$. This construction is the directed graph induced by the ideal/crisp idealized agents.

The construction corresponds respectively to: $C_1=C_2=G_N$ as the search space; $\mu^{(t)}$ as the iterated ideal agent; $\mu^{0,(t)}$ as its safety envelope; $\mathcal{G}_t$ as the geometry induced by the safety envelope.

\begin{figure}[ht!]
    \centering
    \includegraphics[width=.65\linewidth]{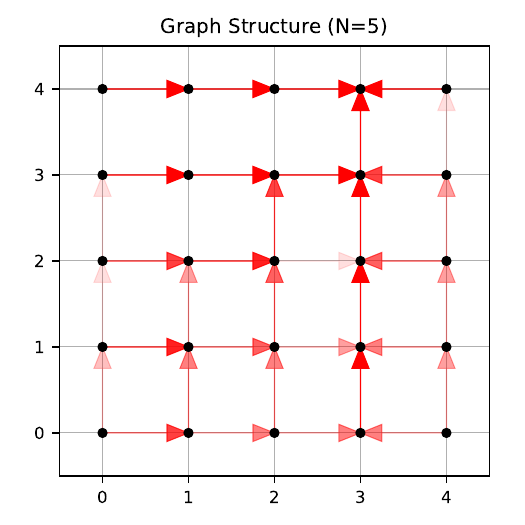}
    \caption{Visualization of $\mathcal{G}_{(3,4)}$ on a $5\times 5$ grid. Red arrows denote reachable directed edges, and transparency encodes the membership on the ideal agent $\mu^{(t)}$. The graph is unidirectional, strictly decreasing the Manhattan distance to the target.}
    \label{fig:N5}
\end{figure}

Experimental setup, model list, grid sizes and target points, and full prompts are provided in Appendix App.~\ref{app:prompt}. Under this setup, we construct the ideal agent $\mu^{(t)}$ for each target $t$, and obtain the crisp agent and the induced directed graph $\mathcal{G}_t$ from its support. Figure~\ref{fig:N5} shows a representative case ($N=5$, $t=(3,4)$) of $\mathcal{G}_t$: under semantic constraints, the graph exhibits a unidirectional structure (strictly decreasing Manhattan distance to the target) with anisotropic preferences over allowed edges, consistent with the finite terms premise in Assumption~\ref{ass:threshold}.

\subsection{IdeaSearch agent}
To further test the observability of Assumption~\ref{ass:threshold} in real complex tasks, we instantiate an iterated agent and its safety envelope in a symbolic regression setting. Specifically, we adopt the IdeaSearchFitter framework proposed by Song et al., configure hyperparameters under "expert mode" and set example\_num to 1 \cite{song2025iteratedagentsymbolicregression}. Under this configuration, the search space $C_1=C_2$ is the numexpr hypothesis space; each state $f$ represents a candidate expression (merging equal expressions(ideas) by string-level equality), and $f\to g$ denotes the next-step update retained by the framework under constraints and scoring filters.

On the PMLB dataset strogatz\_bacres2 \cite{pmlb}, we conduct 10 independent runs, and aggregate 34,651 input–output pairs $f\to g$. To avoid misclassifying occasional random temperature perturbations as input–output edges of an agent constructed by large language models (LLMs), we tally all observed input–output pairs and introduce a threshold parameter $p_0$ for binarization (note: $p_0$ is unrelated to the continuation parameter $p$ in Section 2). Let $n_{\mathrm{in}}(f)$ be the number of occurrences where $f$ appears as the input in all pairs, and let $n_{\mathrm{pair}}(f,g)$ be the number of occurrences of the pair $f\to g$; define the relative frequency $r(f,g):=n_{\mathrm{pair}}(f,g)/n_{\mathrm{in}}(f)$. Based on this, we construct the corresponding crisp agent (safety envelope)
\begin{align}
    \mathcal{T}^{0,(p_0)}_{\mathrm{ideasearch}}(f,g) = \begin{cases}
        0,& n_{\mathrm{in}}(f)\le 1\\
        0,& r(f,g)\le p_0\\
        1,& \text{otherwise}.\\
    \end{cases}
\end{align}
This binarization ensures that an input is included in the envelope only when it has a statistically stable output, thereby filtering out occasional input–output pairs as much as possible.

\begin{figure}[ht!]
    \centering
    \includegraphics[width=1\linewidth]{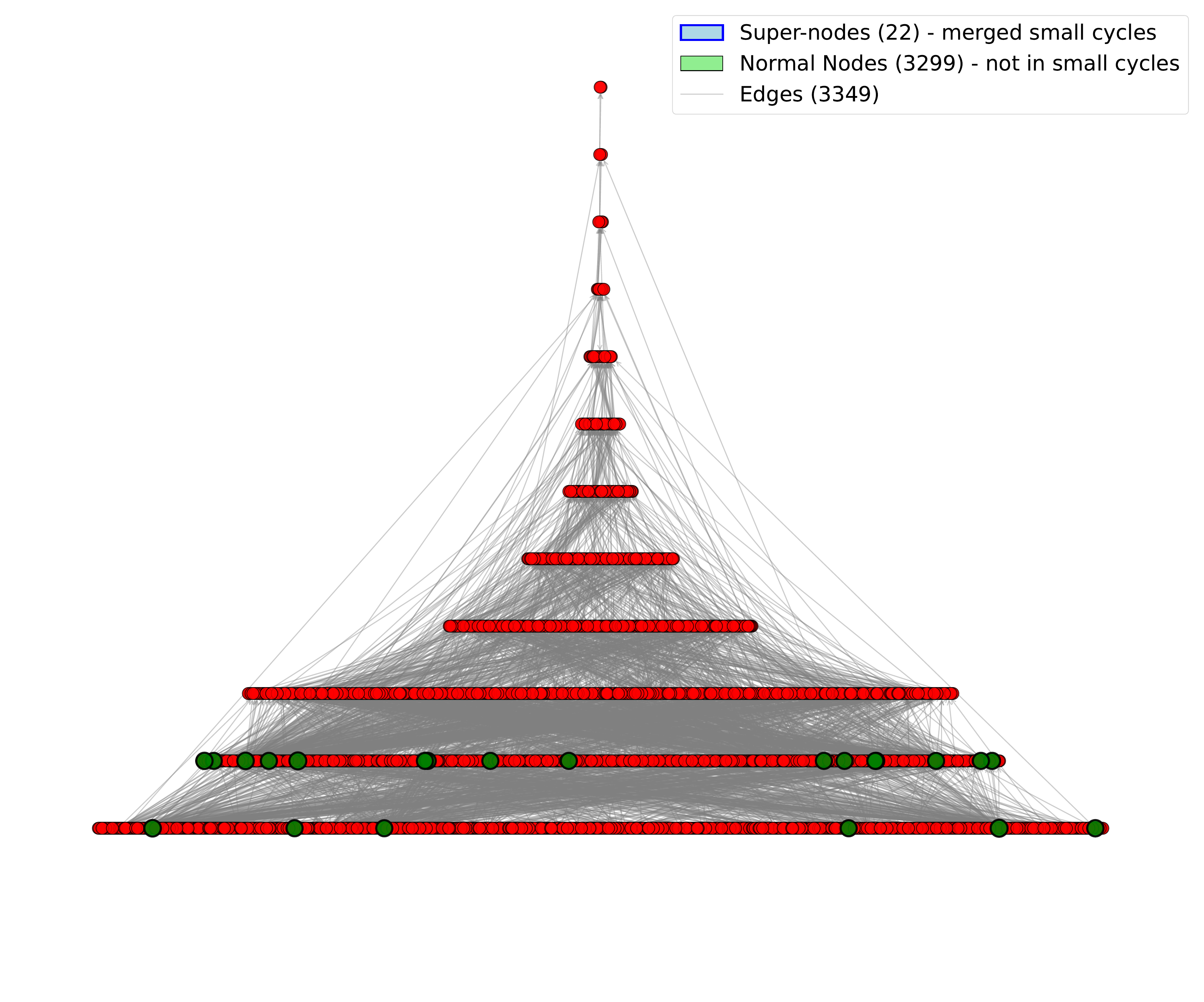}
    \caption{Directed graph after supernode contraction (p0=0.1). Red denotes ordinary nodes, totaling 3299; green denotes supernodes, totaling 22; in sum there are 3349 distinct input–output pairs at the string level. Overall, closed walks are rare, and the graph structure is approximately acyclic.}
    \label{fig:DAG}
\end{figure}

\begin{figure}[ht!]
    \centering
    \includegraphics[width=.7\linewidth]{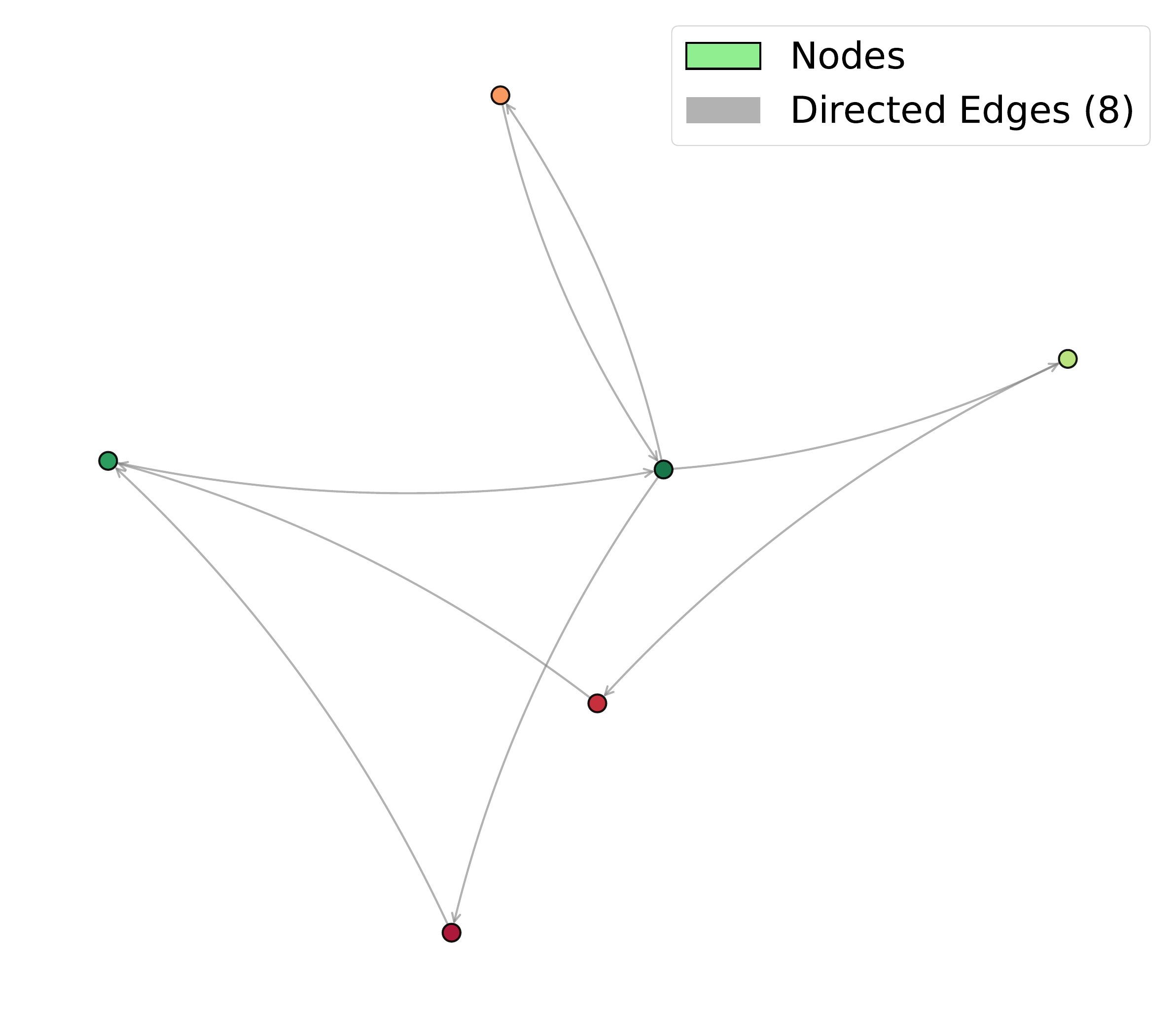}
    \caption{Internal landscape of a supernode (p0=0.1). This supernode contains 6 nodes and 8 internal edges and is strongly connected internally; node color encodes goodness of fit on the dataset (green indicates a lower chi-square goodness-of-fit statistic).}
    \label{fig:inter}
\end{figure}

To test approximate unidirectionality, we contract the few nodes that participate in closed walks into strongly connected components, yielding a supernode graph; Figure~\ref{fig:DAG} shows the directed graph under this contraction, and Figure~\ref{fig:inter} shows the internal landscape of one such supernode. Statistics indicate that when $p_0=0.1$, the total size of all supernodes accounts for about 2\% of all nodes, indicating that closed walks are rare and the induced graph is overall close to acyclic, consistent with the rareness of closed walks in Assumption~\ref{ass:threshold}.

\begin{figure}[ht!]
    \centering
    \includegraphics[width=1\linewidth]{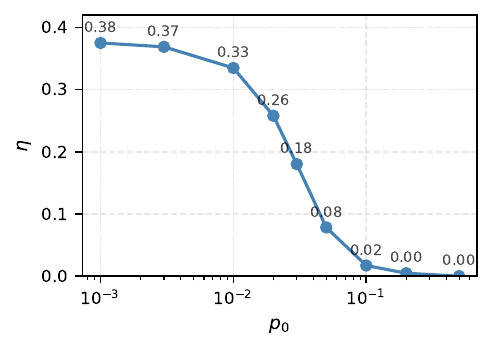}
    \caption{Closed-walk proportion $\eta$ versus threshold $p_0$. $\eta$ is defined as the ratio of the number of nodes contained in supernodes to the total number of nodes; as $p_0$ increases, low-frequency edges are pruned, and $\eta$ rapidly decreases toward zero, indicating the emergence of an approximately unidirectional structure.}
    \label{fig:max_loop}
\end{figure}

Furthermore, we examine the effect of the threshold $p_0$ on the graph structure. Define
\[
\eta(p_0):=\frac{\text{number of nodes contained in supernodes}}{\text{total number of nodes in the graph}}\,,
\]
as an observable measure of the proportion of closed walks. Figure~\ref{fig:max_loop} shows $\eta$ as a function of $p_0$: as $p_0$ increases, low-frequency edges (more likely originating from random temperature perturbations) are systematically pruned; the input–output pairs of the LLM-based reasoning agent gradually dominate the structure, and the number of closed walks drops rapidly. This continuous process exhibits a transition from "random guessing" to "stable reasoning", and macroscopically yields an approximately unidirectional directed graph (approximately acyclic), consistent with the approximate unidirectionality in Assumption~\ref{ass:threshold}.

This instantiation corresponds, at the formal level, to the objects in Section~\ref{sec:formal}: $C_1=C_2$ is the search space of the iterated agent; $\mathcal{T}^{0,(p_0)}_{\mathrm{ideasearch}}$ is the safety envelope obtained via frequency-threshold binarization; the graph induced by the envelope, after contracting closed walks, exhibits an approximately acyclic geometric structure, providing external evidence for the empirical hypothesis of "rare closed walks and rare overly long trajectories".

\section{Conclusion}
This paper proposes a compact formal theory to unify the description and measurement of LLM-assisted iterative search. The core is to represent agents as fuzzy relation operators $\mu$ on inputs and outputs; aggregate the contributions of all reachable paths via the coverage generating function $P_{f,g}(p)$; and characterize reachability difficulty by the critical parameter $p_c(f,g)$ (with coverage index $R_c=1-p_c$). On the graph induced by the crisp agent (safety envelope), the shortest distance $d_0$ and the number of shortest paths $N_{d_0}$ provide a geometric interpretation of the search process. The low-order dominance seen in iterated agents constructed by LLMs supplies a computable, testable, and model-agnostic language for how to measure. The instantiation shows that the safety envelope induced by LLMs yields a unidirectional strcture.

The theory offers testable predictions and quantifiable trade-offs. First, the safety-reachability trade-off can be quantified via $\mu$ and $P_{f,g}(p)$: tightening the safety envelope (reducing reachable edges) decreases path diversity, increases $d_0$, lowers $P_{f,g}(p)$, raises $p_c$, and reduces $R_c$; relaxing constraints has the opposite effect, but must respect safety \cite{irving2018aisafetydebate}. Second, the multiplicative lower bound and the propagation inequality for critical parameters in the presence of intermediate nodes provide possible guidance for constructing intermediate waypoints to reduce overall difficulty \cite{10.1145/1553374.1553380}.
Practically, this theory provides quantitative guidelines for agent design and training on complex tasks. For example, evaluation and training signals can be designed around $p_c$/$R_c$, $d_0$, and $N_{d_0}$ so that reachability difficulty and safety compliance are simultaneous optimization goals; conversely, the theory can guide the design of agents for executing complex tasks: in early stages, prefer models with stricter safety envelopes to shrink the envelope and ensure compliance and stability; once the running epochs approach the reachable limit, gradually introduce looser safety envelopes to increase the coverage index \cite{alphaevolve,song2025iteratedagentsymbolicregression}.

This paper presents an implementable theory; detailed experimental validation is left to future work, including further testing of the effectiveness of these measures and connecting the above indicators to reinforcement learning rewards and training procedures. Overall, by formalizing agents as computable fuzzy relation operators and unifying safety and reachability under the same measurement, the theory serves as a foundational tool for understanding and improving LLM-driven long-horizon search and complex-task agents.

\bibliography{cite}
\bibliographystyle{unsrtnat}

\appendix
\section{Experimental Setup and Prompts}\label{app:prompt}

For reproducibility, this appendix provides detailed experimental setup and prompts.

\subsection{Grid sizes and target points}
\begin{table}[H]
    \centering
    \begin{tabular}{ccc}
         number & $N$ & $t$\\
         \midrule
         1 & 3 & $(1,2)$\\
         2 & 5 & $(3,4)$\\
         3 & 8 & $(6,7)$\\
    \end{tabular}
    \caption{Three grid sizes and corresponding target points $t$.}
    \label{tab:params}
\end{table}

\subsection{Model list and sampling settings}
\paragraph{Model set:} gpt-5-mini, gpt-5, qwen3, qwen-plus, gemini-2.5-flash, deepseek-v3, grok-4, doubao.
\paragraph{Number of samples:} For each input position $f$ under a given target $t$, independently sample $m=5$ times.

\subsection{Prompts}
The prompt used to drive each model to output the next position is as follows:
\begin{lstlisting}
# example input:
# N = 5
# f = (0, 0)
# t = (3, 4)

prompt = f"""
    You are an ant on a {N}x{N} grid. Your current position is {list(f)}, and the target position (food) is {list(t)}.
    You can move up, down, left, or right by one unit, but cannot move outside the grid.
    Based on the current state, decide the next position to move to, and return the result in JSON format with the field "next_position".

    Note:
    - Only choose legal move positions
    - Choose the position that gets you closer to the target
    - Example return format: {{"next_position": [x_g, y_g]}}
    - Write down the json only, no other text
    """
    
messages = [
    {"role": "system", "content": "You are a helpful assistant that helps people find information."},
    {"role": "user", "content": prompt}
]
\end{lstlisting}

\end{document}